\documentclass{article}

\usepackage[nonatbib,final]{nips_2017_arxiv}

\usepackage[utf8]{inputenc} 
\usepackage[T1]{fontenc}    
\usepackage{hyperref}       
\usepackage{url}            
\usepackage{booktabs}       
\usepackage{amsfonts}       
\usepackage{nicefrac}       
\usepackage{microtype}      

\usepackage{amssymb,amsmath,epsfig}
\usepackage{multirow}
\usepackage{booktabs}
\usepackage[caption=false]{subfig}
\usepackage{capt-of}
\usepackage{setspace}
\usepackage{algorithm}
\usepackage{algorithmic}

\title{Early Improving Recurrent Elastic Highway Network}

\author{
  Hyunsin~Park\\
  Department of EE, KAIST\\
  Daejeon, South Korea\\
  \texttt{hs.park@kaist.ac.kr} \\
  \And
  Chang D.~Yoo \\
  Department of EE, KAIST \\
  Daejeon, South Korea \\
  \texttt{cd\_yoo@kaist.ac.kr} \\
}

\begin{document}

\maketitle

\begin{abstract}
To model time-varying nonlinear temporal dynamics in sequential data, a recurrent network capable of varying and adjusting the recurrence depth between input intervals is examined. 
The recurrence depth is extended by several intermediate hidden state units, 
and the weight parameters involved in determining these units are dynamically calculated.
The motivation behind the paper lies on overcoming a deficiency in Recurrent Highway Networks and improving their performances which are currently at the forefront of RNNs: 1) Determining the appropriate number of recurrent depth in RHN for different tasks is a huge burden and just setting it to a large number is computationally wasteful with possible repercussion in terms of performance degradation and high latency. Expanding on the idea of adaptive computation time (ACT), with the use of an elastic gate in the form of a rectified exponentially decreasing function taking on as arguments as previous hidden state and input, the proposed model is able to evaluate the appropriate recurrent depth for each input. 
The rectified gating function enables the most significant intermediate hidden state updates to come early such that significant performance gain is achieved early.  
2) Updating the weights from that of previous intermediate layer offers a richer representation than the use of shared weights across all intermediate recurrence layers. The weight update procedure is just an expansion of the idea underlying hypernetworks.
To substantiate the effectiveness of the proposed network, we conducted three experiments: regression on synthetic data, human activity recognition, and language modeling on the Penn Treebank dataset. 
The proposed networks showed better performance than other state-of-the-art recurrent networks in all three experiments.
\end{abstract}

\section{Introduction}

Recurrent Neural Networks (RNNs) have been successfully applied to a diverse array of tasks, such as speech recognition \cite{graves2014towards}, language modeling \cite{mikolov2012statistical}, machine translation \cite{cho2014properties}, music generation \cite{boulanger2012modeling}, image description \cite{xu2015show}, video description \cite{yao2015describing}.
In these tasks, the non-linear transition of the internal state in the RNNs represented the temporal dynamic behavior of the sequence. This paper investigates the effectiveness of elastically varying the intermediate recurrence depth of the RNN in modeling the time-varying temporal dynamics of the sequential input data. The intermediate recurrence depth at each time interval will depend on the input and hidden state at a corresponding time interval.

While deeper networks have been known to be more efficient in representing the input-output relationship compared to shallow networks, they require more data to train and are far more susceptible to the vanishing and exploding gradient problem. 
As a measure to safeguard against such problem when constructing a very deep network, ResNet \cite{he2016deep} and Highway Networks \cite{srivastava2015training} were proposed. Residual units in these models provide learning stability in building a very deep model.
In creating temporally deep RNNs, Long Short-Term Memory (LSTM, \cite{hochreiter1997long}), Gated Recurrent Unit (GRU, \cite{cho2014properties}), and Recurrent Highway Network (RHN, \cite{zilly2016recurrent}) were introduced to both find learning stability and model long-range temporal dependency.

One factor that affects the computational time is the depth (or the number of layers) of the network.
Given a training dataset, to get good performance, various models with different depth are empirically compared, and this can be a very time-consuming procedure if the dataset is large.

Recently, two studies on determining appropriate depth of a network have been considered.
A general concept of time-dependent adaptive computation time (ACT) is introduced into RNN in \cite{graves2016adaptive}. The algorithm learns the number of computational steps to take between receiving an input and emitting an output. Here, a sigmoidal halting unit determines the probability for continuing the computation.
In Spatially Adaptive Computation Time \cite{figurnov2016spatially}, the number of executed layers for the regions of the image is dynamically adjusted.

To model time-varying nonlinear temporal dynamics in sequential data, a recurrent network capable of varying and adjusting the recurrence depth between input intervals is examined. 
The recurrence depth is extended by several intermediate hidden state units, 
and the weight parameters involved in determining these units are dynamically calculated.
The motivation behind the paper lies on overcoming a deficiency in Recurrent Highway Networks and improving their performances which are currently at the forefront of RNNs: 1) Determining the appropriate number of recurrent depth in RHN for different tasks is a huge burden and just setting it to a large number is computationally wasteful with possible repercussion in terms of performance degradation and high latency. Expanding on the idea recently proposed in ACT \cite{graves2016adaptive}, with the use of an elastic gate in the form of a rectified exponentially decreasing function taking on as arguments as previous hidden state and input, the proposed model is able to evaluate the appropriate recurrent depth for each input. 
The rectified gating function enables the most significant intermediate hidden state updates to come early such that significant performance gain is achieved early.  
2) Updating the weights from that of previous intermediate layer offers a richer representation than the use of shared weights across all intermediate recurrence layers. The weight update procedure is just an expansion of the idea underlying hypernetworks \cite{ha2016hypernetworks}.
This structure is referred to as Early Improving Recurrent Elastic Highway Network (EI-REHN).

This paper is organized as follows. 
Section 2 describes background information for the proposed method.
Section 3 presents the proposed method of EI-REHN.
Section 4 shows the experimental results,
and finally, Section 5 concludes the paper.

\section{Background}

Consider a hidden state transition model $\mathcal{S}$ of a recurrent network,
\begin{eqnarray}
\mathbf{h}_t = \mathcal{S}(\mathbf{h}_{t-1}, \mathbf{x}_t),
\end{eqnarray}
where $\mathbf{x}_t \in \mathbb{R}^{D_x}$ and $\mathbf{h}_{t-1} \in \mathbb{R}^{D_h}$ are the input vector at time step $t$ and hidden state vector at time step $t-1$, respectively.

The hidden state transition model of the standard RNN can be represented as follows,
\begin{eqnarray}
\mathcal{S}_{RNN}(\mathbf{h}_{t-1}, \mathbf{x}_t) = \mbox{tanh}\left( \mathbf{W}_R 
\left[\begin{array}{c} \mathbf{h}_{t-1} \\ \mathbf{x}_t \\ 1 \end{array}\right] \right),
\end{eqnarray}
where $\mathbf{W}_{R} \in \mathbb{R}^{D_h \times (D_h + D_x + 1)}$ and $\mbox{tanh}$ are a weight matrix including bias and element-wise hyperbolic tangent function, respectively.

On the other hand, the hidden state transition model of an LSTM is as follows,
\begin{eqnarray}
\mathcal{S}_{LSTM}(\mathbf{h}_{t-1}, \mathbf{c}_{t-1}, \mathbf{x}_t) &=& \mathbf{o}_t \otimes \mbox{tanh}(\mathbf{c}_t),
\end{eqnarray}
where
\begin{eqnarray}
\mathbf{c}_t &=& \mathbf{f}_t \otimes \mathbf{c}_{t-1} + \mathbf{i}_t \otimes \mathbf{a}_t,\\
\left( \begin{array}{c} \mathbf{a}_t \\ \mathbf{i}_t \\ \mathbf{f}_t \\ \mathbf{o}_t \end{array}\right) &=& 
\left( \begin{array}{c} \mbox{tanh} \\ \mbox{sigm} \\ \mbox{sigm} \\ \mbox{sigm} \end{array}\right)
\mathbf{W}_L \left[ \begin{array}{c} \mathbf{h}_{t-1} \\ \mathbf{x}_t \\ 1 \end{array}\right]. 
\end{eqnarray}
Here, $\mathbf{a}_t \in \mathbb{R}^{D_h}$, $\mathbf{i}_t \in \mathbb{R}^{D_h}$, $\mathbf{f}_t \in \mathbb{R}^{D_h}$, $\mathbf{o}_t \in \mathbb{R}^{D_h}$, $\mathbf{c}_t \in \mathbb{R}^{D_h}$, $\mathbf{W}_L \in \mathbb{R}^{4D_h \times (D_h + D_x + 1)}$, $\mbox{sigm}$, and $\otimes$ are 
cell proposal, input gate, forget gate, output gate, cell state, weight matrix, element-wise sigmoid function, and element-wise multiplication, respectively.

The hidden state transition model of RHN with a fixed recurrence depth is as follows,
\begin{eqnarray}
\mathcal{S}_{RHN}(\mathbf{h}_{t-1}, \mathbf{x}_t) &=& \mathbf{h}_t^R,
\end{eqnarray}
where
\begin{eqnarray}
\mathbf{h}_t^0 &=& \mathbf{h}_{t-1}  \nonumber \\
\mathbf{h}_t^r &=& \mathbf{t}_t^r \otimes \mathbf{s}_t^r + \mathbf{c}_t^r \otimes \mathbf{h}_t^{r-1}, \\
\left( \begin{array}{c} \mathbf{s}_t^r \\ \mathbf{t}_t^r \\ \mathbf{c}_t^r \end{array}\right) &=& 
\left( \begin{array}{c} \mbox{tanh} \\ \mbox{sigm} \\ \mbox{sigm} \end{array}\right)
\mathbf{W}_H^r \left[ \begin{array}{c} \mathbf{h}_{t}^{r-1} \\ \mathbb{I}_{\{r=1\}} \mathbf{x}_t \\ 1 \end{array}\right]. 
\end{eqnarray}
Here, $\mathbf{s}_t^r \in \mathbb{R}^{D_h}$, $\mathbf{t}_t^r \in \mathbb{R}^{D_h}$, $\mathbf{c}_t^r \in \mathbb{R}^{D_h}$, and $\mathbf{W}_H^r \in \mathbb{R}^{3D_h \times (D_h + D_x + 1)}$, are residual component, transform gate, carry gate, and weight matrix, respectively.

\section{Early Improving Recurrent Elastic Highway Network}

In this section, Early Improving Recurrent Elastic Highway Network (EI-REHN) which is a recurrent network capable of varying and adjusting the recurrence depth between input intervals is introduced.
The intermediate state transition from the $(r-1)$-th intermediate recurrence layer to the $r$-th layer is given as 
\begin{eqnarray}
\mathbf{h}_t^{r} &=& \mathbf{g}_t^r \otimes \mathbf{s}_t^r + (\mathbf{1} - \mathbf{g}_t^r) \otimes \mathbf{h}_t^{r-1}, 
\label{eq-rhn}
\end{eqnarray}
where $\mathbf{g}_t^i$ and $\mathbf{s}_t^r$ are a gating function for adaptive recurrence depth and residual component, respectively.
The gating function is designed to exponentially decrease as the intermediate recurrence layer increases.
Consequently, intermediate recurrence state transition halts when the gating function reaches zero.
The exponentially decreasing gating function enables the most significant intermediate hidden state updates to come early such that significant performance gain is achieved early.
In order to reduce the number of parameters, a recurrence relationship is formulated between parameters of adjacent layers such that only the significant parameters are updated. 
And the weight parameters for all the intermediate recurrence layers are calculated based on a hypernetwork \cite{ha2016hypernetworks} that is a sub-network to generate the weights for another network.

\begin{figure}[t]
\begin{center}
\includegraphics[width=0.8\textwidth]{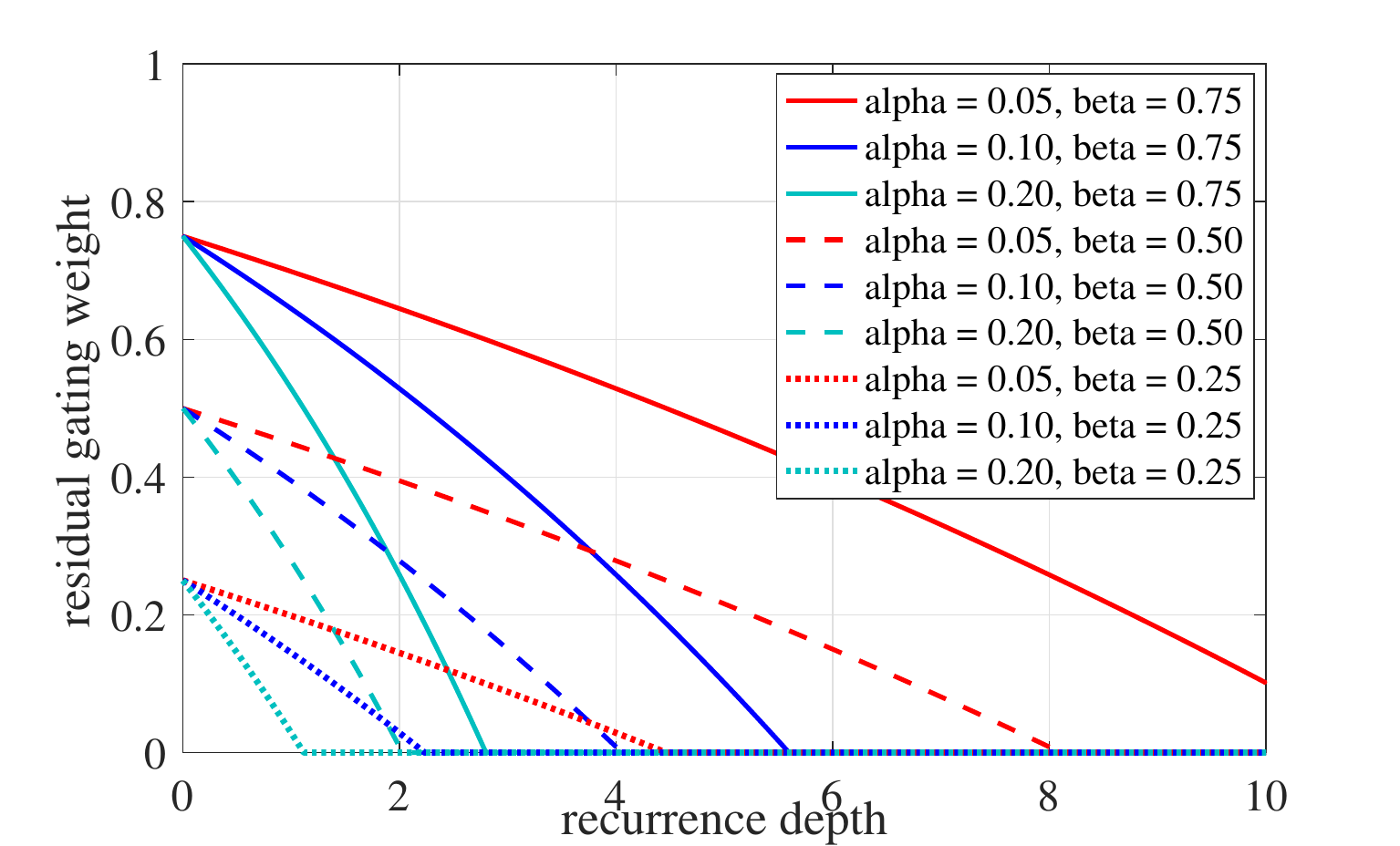}
\end{center}
\caption{Examples of global upper-bounds to elastic gating functions.}
\label{fig-rgf}
\end{figure}

\subsection{Adaptive recurrence depth}

A gating function to adaptively determine the intermediate recurrence depth depending on $\mathbf{x}_t$ and $\mathbf{h}_{t-1}$ is given as follows:
\begin{eqnarray}
\mathbf{g}_t^r &=& \mathbf{d}_t^r \otimes \hat{\mathbf{g}}_t^r,
\end{eqnarray}
where $\mathbf{d}_t^r$ and $\hat{\mathbf{g}}_t^r$ are elastic gating and residual gating, respectively.
The elastic gating function is obtained as follows,
\begin{eqnarray}
\mathbf{d}_t^r &=& \mbox{max} (\boldsymbol{\beta} + e^{\boldsymbol{\alpha}} - e^{(\boldsymbol{\alpha} + \boldsymbol{\alpha}_t) r}, 0), \\
\boldsymbol{\beta} &=& \mbox{sigm}(\hat{\boldsymbol{\beta}}), \\
\boldsymbol{\alpha} &=& \mbox{softplus}(\hat{\boldsymbol{\alpha}}), \\
\boldsymbol{\alpha}_t &=& \mbox{sigm} \left( \mathbf{W}_a 
\left[\begin{array}{c} \mathbf{h}_{t-1} \\ \mathbf{x}_t \\ 1 \end{array} \right] \right),
\end{eqnarray} 
where $\boldsymbol{\beta}$, $\boldsymbol{\alpha}$, $\boldsymbol{\alpha}_t$, and $\mathbf{W}_a \in \mathbb{R}^{D_h \times (D_h + D_x + 1)}$ are initial gating bias, global decreasing rate, local decreasing rate, and weight matrix for residual gating function, respectively. 
The model parameters of $\hat{\boldsymbol{\alpha}}$ and $\hat{\boldsymbol{\beta}}$ are estimated when training, while the local decreasing rate, $\boldsymbol{\alpha}_t$, is time-dependent and activated when forward propagation. 
It makes the model to have adaptive recurrence depth at every time step.

At time $t$, adaptive recurrence depth $R_t$ is determined as the maximum depth that satisfies $ ||\mathbf{g}_t^r||_1 > 0$. This means that the recurrence layer repeats until all hidden units have zero gating activations.
Based on $\boldsymbol{\alpha}$ and $\boldsymbol{\beta}$, upper-bound to $R_t$ for all the time steps is 
\begin{eqnarray}
R_t = \max_{||\mathbf{g}_t^r||_1 > 0} r \le \max_i \left\lfloor \frac{1}{\alpha_i} \log(\beta_i+e^{\alpha_i}) \right\rfloor.
\end{eqnarray}
Figure \ref{fig-rgf} shows examples of elastic gating functions for a hidden unit when $\boldsymbol{\alpha}_t^r = 0$. The elastic gating function is exponentially decreasing with respect to recurrence depth. Smaller value of $\alpha$ and bigger value of $\beta$ give longer adaptive recurrence depth.

If an element of $\mathbf{d}_t^{r=1}$ is zero, the corresponding hidden unit is not updated at time step $t$ and the hidden information from the previous time step is passed to the next time step due to the exactly zero gating activation.

\subsection{Dynamic weight matrix}

At each intermediate recurrence layer, the residual component is calculated as follows,
\begin{eqnarray}
\mathbf{s}_t^r = \mbox{tanh}(\mathbf{W}_x \mathbf{x}_t \mathbb{I}_{r=1} + \mathbf{W}_s^r \mathbf{h}_t^{r-1} + \mathbf{b}_s^r), 
\end{eqnarray}
where $\mathbf{W}_x \in \mathbb{R}^{D_h \times D_x}$, $\mathbf{W}_s^r \in \mathbb{R}^{D_h \times D_h}$, and  $\mathbf{b}_s^r \in \mathbb{R}^{D_h}$ are input-to-residual weight matrix, hidden-to-residual weight matrix, and residual bias, respectively. The residual gating $\hat{\mathbf{g}}_t^r$ is also calculated in similar manner by replacing 'tanh' with 'sigm'. Here, $\mathbf{W}_s^r$ and $\mathbf{b}^r$ are dependent on the recurrence layer $r$. In the proposed model, the recurrence depth is time-varying and can be very deep. 
Learning all weight matrices for all possible depth is impractical. 
One solution is to use a shared weight matrix. 
However, this constraint reduces the ability of feature representation. To solve this problem, we propose a dynamic weight matrix utilizing the concept of hypernetworks \cite{ha2016hypernetworks}.

At $t$ time step and $r$ intermediate recurrence layer, for the calculations of the residual component and residual gating, we use a dynamic weight matrix that is different for each calculation and defined as follows.
\begin{eqnarray}
\mathbf{W}_t^r = \mathbf{W}_t^{r-1} + \Delta \mathbf{W}_t^r,
\end{eqnarray}
where, $\Delta \mathbf{W}_t^r$ is obtained by hypernetwork that is a simple RNN with small number of hidden units compared with the number of main hidden units. 
The hypernetwork takes input as the previous residual component $\mathbf{s}_t^{r-1}$  and residual gating $\hat{\mathbf{g}}_t^{r-1}$ and calculates the hidden hypernetwork state $\mathbf{z}_t^r \in \mathbb{R}^{D_z}$,
\begin{eqnarray}
\mathbf{z}_t^{r} &=& \tanh ( \mathbf{W}_{zh} \mathbf{s}_t^{r-1} + \mathbf{W}_{zg} \hat{\mathbf{g}}_t^{r-1} +  \mathbf{W}_z \mathbf{z}_t^{r-1} + \mathbf{b}_z ),
\end{eqnarray}
where $\mathbf{W}_{zh} \in \mathbb{R}^{D_z \times D_h}$, $\mathbf{W}_{zg} \in \mathbb{R}^{D_z \times D_h}$, $\mathbf{W}_z \in \mathbb{R}^{D_z \times D_z}$, and $\mathbf{b}_z \in \mathbb{R}^{D_z}$ are model parameters of the hypernetwork.
From the hidden state of the hypernetwork, we calculate a diagonal $\Delta \mathbf{W}_t^r$ as follows,
\begin{eqnarray}
\mathbf{w}_{t}^{r} &=& \mathbf{P} \mathbf{z}_t^{r}, \\
\Delta \mathbf{W}_t^r &=&  \mbox{diag}(\mathbf{w}_t^r),
\end{eqnarray}
where $\mathbf{P} \in \mathbb{R}^{D_h \times D_z}$ is a projection matrix.
With the $\Delta \mathbf{W}_t^r$, we consider gated weight update as follows.
\begin{eqnarray}
\bar{\mathbf{g}}_t^r &=& \mbox{sigm}(\bar{\mathbf{P}} \mathbf{z}_t^r + \bar{\mathbf{b}}), \\
\mathbf{s}_t^r &=& \mbox{tanh}( \bar{\mathbf{g}}_t^r \otimes \mathbf{W}_t^{r-1} \mathbf{h}_t^{r-1} + (1-\bar{\mathbf{g}}_t^r) \otimes \Delta\mathbf{W}_t^{r} \mathbf{h}_t^{r-1} + \mathbf{W}_x\mathbf{x}_t\mathbb{I}_{r=1} + \mathbf{b}).
\end{eqnarray}

We tried various ways of updating the parameters (e.g. weight scaling of hypernetworks in \cite{ha2016hypernetworks}, weight matrix decomposition proposed by Jurgen Schmidhuber in \cite{schmidhuber1992learning2}, etc.) and found diagonal update to be the most effective. 

\begin{figure}[t]
\begin{center}
\includegraphics[width=0.9\textwidth]{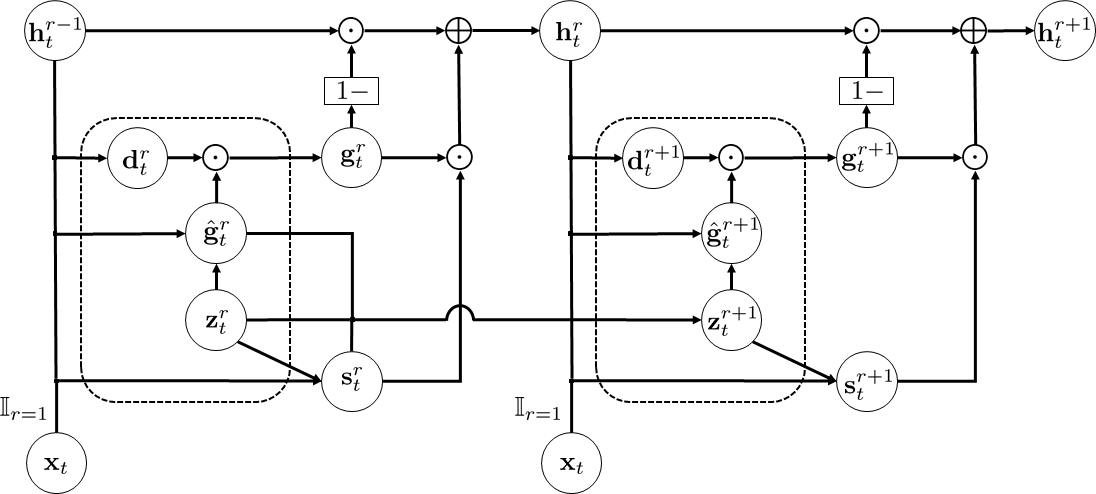}
\end{center}
\caption{Schematic showing computation of EI-REHN at $t$ time step from $r-1$ to $r+1$ recurrence layer. The $\odot$, $\oplus$, and '$1-$' mean element-wise product, sum, and one minus the input, respectively. Dotted round square represents the extended part from RHN. For clarity, some nodes and connections are omitted.}
\label{fig-adrhn}
\end{figure}

Finally, the procedure of the state transition with adaptive recurrence depth is described in Algorithm \ref{alg:rehn}.
Computation graph of the proposed network is shown in Figure \ref{fig-adrhn}.

\begin{figure}[ht]
\begin{minipage}[t]{.475\linewidth}
\begin{algorithm}[H]\small
   \setstretch{1.14}
   \caption{State transition with adaptive recurrence depth}
   \label{alg:rehn}
   \algsetup{indent=.5em}
\begin{algorithmic}
   \STATE {\bfseries Input:} $\mathbf{x}_t$, $\mathbf{h}_{t-1}$, $R_{max}$
   \STATE $r=0, R_t = 0, \mathbf{h}_t^0 = \mathbf{h}_{t-1}$
   \STATE Calculate decreasing rate $\boldsymbol{\alpha}_t$ in Eq.(14)   
   \WHILE{true}
   \STATE $r = r + 1$
   \STATE Update hypernetwork state $\mathbf{z}_t^r$ in Eq.(18)
   \STATE Calculate residual component $\mathbf{s}_t^r$ in Eq.(22)
   \STATE Get gating function $\mathbf{g}_t^r$ in Eq.(10)
   \IF{$||\mathbf{g}_t^r||_1 > 0$ and $r \le R_{max}$}
   \STATE $R_t = r$
   \STATE Update intermediate hidden state $\mathbf{h}_t^r$ in Eq.(9)   
   \ELSE 
   \STATE {\bfseries break}
   \ENDIF   
   \ENDWHILE
   \STATE {\bfseries Output:} $\mathbf{h}_t = \mathbf{h}_t^{R_t}$
\end{algorithmic}
\end{algorithm}
\end{minipage}
~~~~~
\begin{minipage}[t]{.475\linewidth}
\begin{algorithm}[H]\small
\caption{Synthetic data generation}
\label{alg:syn}
\algsetup{indent=.4em}
\begin{algorithmic}
\STATE {\bfseries Input:} $N, T, R_{max}, \theta$
\STATE Make an empty synthetic dataset $\mathcal{X} = \{ \}$
\FOR{$n = 1$ to $N$}
\STATE Draw $\mathbf{h}_0 \sim U[-1,1]$
\FOR{$t = 1$ to $T$}
\STATE $\mathbf{h}_t^{0} = \mathbf{h}_{t-1}$
\STATE $R_t = \mbox{round}((R_{max} -1) * ||\mathbf{h}_{t-1}||_2^2) + 1$
\FOR{$r = 1$ to $R_t$}
\STATE $\mathbf{n}_r \sim \mathcal{N}(\mathbf{0}, 0.1 \times \mathbf{I})$
\STATE $\mathbf{h}_t^r = \mbox{tanh} \left(\left[ \begin{array}{rr} 
\cos (\theta) & -\sin (\theta) \\ \sin (\theta) & \cos (\theta)
\end{array}\right] \mathbf{h}_t^{r-1} + \mathbf{n}_r \right)$
\ENDFOR
\STATE $\mathbf{h}_t = \mathbf{h}_t^{R_t}$
\STATE $\mathbf{x}_t = \frac{R_t}{R_{max}} \times  \left[ \begin{array}{c} \mbox{tanh}(h_t(1) + h_t(2)) \\ \mbox{tanh}(h_t(1) - h_t(2)) \end{array} \right]$
\ENDFOR
\STATE Add sequence $[\mathbf{x}_1, \cdots, \mathbf{x}_T]$ to $\mathcal{X}$
\ENDFOR
\STATE {\bfseries Output:} $\mathcal{X}$
\end{algorithmic}
\end{algorithm}
\end{minipage}
\end{figure}

\section{Experiments}

\subsection{Synthetic data}

This section aims to show the effectiveness of the adaptive recurrence depth on a synthetic dataset by comparing with other recurrent networks.
To this end, we first constructed a synthetic dataset for sequential regression task.

The considered task is defined as predicting two-dimensional real vector of next step after 
observing real vector sequence up to the current step.
The synthetic dataset is generated as described in Algorithm \ref{alg:syn}.
The inputs, $N, T, R_{max}, \theta$, are the number of samples, sequence length, maximum recurrence depth, and affine transform parameter, respectively. We set the inputs as $(N, T, R_{max}, \theta) = (10000, 21, 10, \pi/6)$. The synthetic dataset is divided into the training set, validation set, and test set with  the size of $8000$, $1000$, and $1000$, respectively.
For all the experiments in this paper, Tensorflow toolkit \cite{tensorflow2015-whitepaper} was used. For training the network, Adam optimizer \cite{kingma2014adam} was adopted with $20$ mini-batch size, $100$ epochs, and $0.01$ learning rate. Each model was trained five times with different initial conditions.

\begin{figure}[ht]
\begin{minipage}[t]{.475\linewidth}
\captionof{table}{Sequence regression results on the synthetic dataset.}
\label{tab-syn}
\begin{center}
\begin{tabular}{lcccc}
\toprule
Model & $D_h$ & \# of param. & MSE ($10^{-3})$\\
\midrule
\multirow{3}{*}{RNN}	& 20  & 502 & 1.01 $\pm$ 0.07 \\
						& 30  & 1052 & 0.97 $\pm$ 0.04 \\
						& 40  & 1802 & 1.02 $\pm$ 0.11 \\ \hline
\multirow{3}{*}{LSTM}	& 10  & 542 &  0.76 $\pm$ 0.01 \\ 
						& 15  & 1112 &  0.76 $\pm$ 0.01 \\ 
						& 20  & 1882 & 0.72 $\pm$ 0.01 \\ \hline
\multirow{3}{*}{RHN}	& 10  & 1162 & 0.58 $\pm$ 0.03\\
 						& 15  & 2492 & 0.54 $\pm$ 0.02 \\
		 				& 20  & 4322 & 0.53 $\pm$ 0.02 \\ \hline
\multirow{3}{*}{EI-REHN}	& 10  & 807  & 0.55 $\pm$ 0.02 \\
			 			& 15  & 1732 & 0.50 $\pm$ 0.01 \\
						& 20  & 2862 & $\bold{0.47 \pm 0.01}$ \\
\bottomrule
\end{tabular}
\end{center}
\end{minipage}
~~~~
\begin{minipage}[t]{.475\linewidth}
\captionof{table}{Human action recognition results on the HAR dataset.}
\label{tab-har}
\begin{center}
\begin{tabular}{lcccc}
\toprule
Model & $D_h (R)$ & \# of param. & Accuracy [\%]\\
\midrule
\multirow{3}{*}{RNN}	& 32&  3924  & 82.26 $\pm$ 2.57 \\ 
 						& 48&  8214  & 81.31 $\pm$ 5.53 \\ 
 						& 64&  14022 & 81.11 $\pm$ 4.16 \\ \hline 
\multirow{3}{*}{LSTM} 	& 16&  4048  & 90.88 $\pm$ 0.96 \\ 
						& 24&  8358  & 91.42 $\pm$ 1.06 \\ 
 						& 32&  14214 & 92.12 $\pm$ 0.40 \\  \hline
\multirow{3}{*}{RHN}   	& 32(1)&  7366 	& 90.58 $\pm$ 0.95  \\ 
  						& 32(2)&  11590 & 89.06 $\pm$ 0.61  \\ 
						& 32(3)&  15814 & 83.06 $\pm$ 4.62  \\ \hline
\multirow{3}{*}{EI-REHN} & 32&  8102 	& 91.50 $\pm$ 0.51 \\
						& 40&  12126 	& 91.84 $\pm$ 0.63 \\
						& 48&  16950 	& $\bold{92.48 \pm 0.61}$ \\
\bottomrule
\end{tabular}
\end{center}
\end{minipage}
\end{figure}

To confirm the effectiveness of the elastic gating, we compared 1) an RHN with shared parameters for the intermediate recurrent layers (SRHN), with 2) an RHN with shared parameters and elastic gate (SREHN) for the synthetic dataset. The hidden dimension was set to 20. SREHN showed 0.54 MSE, Now, performances by varying the recurrent depth of SRHN from 1 to 6 produced 0.81, 0.63, 0.58, 0.55, 0.55, and 0.54 MSE. From the results, we can say that the elastic gate helps in finding a recurrent depth that performs comparable to the best RHN with fixed recurrent depth. And the proposed model EI-REHN which is a model that the dynamic weight matrix is added to SREHN showed 0.47 MSE. From this result, we can say that the dynamic weight matrix gives rise to performance gain from 0.54 to 0.47.

We compare the proposed model with RNN, LSTM and RHN on the synthetic dataset by varying model structure parameters such as hidden dimension $D_h$ and recurrence depth $R$.
Table \ref{tab-syn} shows the MSE results with the number of parameters corresponding to each model. 
As shown in Table \ref{tab-syn}, RNN shows the worst performance.
The proposed model shows better MSE performance with a smaller size model than other models.

\subsection{Human activity recognition}

In this subsection, we describe sequence classification experiments by using {\it Human Activity Recognition Using Smartphones Data Set} (HAR) \cite{anguita2013public}.
In the HAR dataset, 30 persons performed six activities (WALKING, WALKING\_UPSTAIRS, WALKING\_DOWNSTAIRS, SITTING, STANDING, LAYING) wearing a smartphone on the waist. 3-axial linear acceleration and 3-axial angular velocity at are captured by using its embedded accelerometer and gyroscope. The dataset is divided into 7352 training sequences and 2947 test sequences.

For the human activity recognition on the HAR dataset, the input sequence is modeled by RNN, LSTM, RHN, and EI-REHN. 
Two-layer recurrent networks are used for this experiments. 
Six-class softmax layer is added to the last hidden units in the top hidden sequence for classification. 
To train the models, cross-entropy loss and Adam optimizer with $200$ mini-batch size, $100$ epochs, and $0.0025$ learning rate are used.

Table \ref{tab-har} shows the sequence classification results. 
In sequence classification, it is important that the last hidden units in recurrent neural networks have to contain information about the whole sequence. In the RHN case, as the recurrence depth increases, the performance decreases. 
We believe that the performance degradation of the RHNs from shallow recurrence depth occurs since the intermediate state transitions vanish the early information of sequence contained in the hidden units in spite of the short-cut path in the RHNs. 
Nonetheless, the proposed model shows better performance than other models in terms of classification accuracy.

\subsection{Language modeling}

\begin{figure}[ht]
\begin{center}
\includegraphics[width=1.0\textwidth]{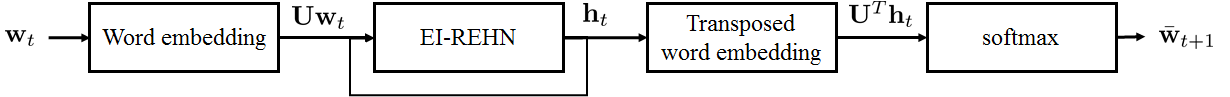}
\end{center}
\caption{Model architecture for language modeling using EI-REHN. The model takes the words in a sequence one by one and predict the next word. Word embedding weight matrix is reused for the output layer.}
\label{fig-lm_rehn}
\end{figure}

In this subsection, word level language modeling on the Penn TreeBank dataset \cite{marcus1993building} is described. 
The model architecture is represented in Figure \ref{fig-lm_rehn}.
For vocabulary of size $C$, the one-hot vector is used to represent the word input at time $t$, $\mathbf{w}_t \in \mathbb{R}^C$ and word embedding vector is obtained by $\mathbf{U} \mathbf{w}_t$, where  $\mathbf{U} \in \mathbb{R}^{H \times C}$. The word embedding vector is passed into EI-REHN. 
The hidden activation of EI-REHN, $\mathbf{h}_t$, is multiplied by the transposed word embedding matrix, $\mathbf{U}^T$ based on \cite{inan2016tying, press2016using}.
Finally, softmax layer is applied to calculate the next word prediction $\hat{\mathbf{w}}_{t+1}$.

\begin{table}[ht]
\caption{\label{tab-ptb1}{Language modeling results on the Penn Treebank dataset.}}
\begin{center}
\begin{tabular}{l|*{2}{cccc|}}
\toprule
 & \multicolumn{4}{c|}{RHN} & \multicolumn{4}{c|}{EI-REHN} \\
\midrule
Recurrence depth & 2 & 3 & 4 & 5 & 2 & 3 & 4 & 5 \\
\# of param. (M) & 20.7 & 23.5 & 26.4 & 29.3 & \multicolumn{4}{c|}{25.3}  \\
Perplexity & 73.4 & 69.8 & 68.3 & 67.6 & 69.2 & 67.4 & $\bold{66.2}$ & 66.8 \\
\bottomrule
\end{tabular}
\end{center}
\end{table}
In Table \ref{tab-ptb1}, we compare EI-REHN with RHN under the fixed number of hidden units. 
For the parameter optimization, we follow the experimental settings described in \cite{zilly2016recurrent}. 
We just replace the `RHNCell' class in the code of \cite{zilly2016recurrent} with an 'EIREHNCell' class that is an implementation of the proposed model.
The recurrence depth for RHN is varied from 2 to 5 while the maximum recurrent depth is set from 2 to 5.
The hidden state size $D_h$ and the hypernetwork state size $D_z$ are set to $1200$ and $600$, respectively.
The number of parameters of RHN increases as the recurrence depth increases, while the number of parameters of EI-REHN doesn't. The EI-REHN with the maximum recurrence depth of 4 shows the best performance in this experiments.
This result supports that the elastic gating of EI-REHN makes the model to fit in early recurrence depths compared with RHN.

\begin{table}[ht]
\caption{\label{tab-ptb}{Comparison of recent word-level language models on the Penn Treebank dataset.}}
\begin{center}
\begin{tabular}{lccc}
\toprule
Model & Size & Val. & Test \\
\midrule
LSTM \cite{zaremba2014recurrent} & 66M & 82.2 & 78.4 \\
Pointer Sentinel networks \cite{merity2016pointer} & 21M & 72.4 & 70.9 \\
Ensemble of LSTMs \cite{zaremba2014recurrent} & - & 71.9 & 68.7 \\
RHN \cite{zilly2016recurrent} & 24M & 68.1 & 66.0 \\
\midrule
EI-REHN-1000D & 19M & 71.9 & 68.7 \\
EI-REHN-1200D & 25M & 70.0 & 66.2 \\
\bottomrule
\end{tabular}
\end{center}
\end{table}
In Table \ref{tab-ptb}, we compare our models ($D_h = 1000, 1200$) with the basic LSTM \cite{zaremba2014recurrent}, Pointer Sentinel networks \cite{merity2016pointer}, Ensemble of LSTMs \cite{zaremba2014recurrent} and RHN \cite{zilly2016recurrent}. From the results, the proposed model shows comparable performance with RHN and better than the other models in terms of perplexity.
\textit{The reported result of RHN was with ten recurrence depth, whereas the proposed model reached the maximum performance with four recurrence depth.}

\section{Conclusion}
To model time-varying nonlinear temporal dynamics in sequential data, a recurrent network capable of varying and adjusting the recurrence depth between input intervals was examined. 
By incorporating into the recurrent network that combines a shortcut path with a residual path with a rectified residual gating function which is best described as a rectified exponentially decreasing function, the network is capable of having varying recurrence depth.
Moreover, we propose dynamic weight matrix construction for recurrence layers. 
This capability extends the capacity of existing recurrent network. 
To substantiate the effectiveness of the proposed network, 
we conducted three experiments that are a regression on the synthetic data, human activity recognition, and language modeling on the Penn Treebank dataset. 
The proposed networks showed better performance than other state-of-the-art recurrent networks in all three experiments.

\small

\bibliographystyle{IEEEtran}
\bibliography{nips_2017_hspark.bbl}

\end{document}